\documentclass[journal]{IEEEtran1}
\usepackage{times}  %Required
\usepackage{helvet}  %Required
\usepackage{courier}  %Required
\usepackage{url}  %Required
\usepackage{graphicx}  %Required
\usepackage{amssymb}
\usepackage[numbers]{natbib}
\usepackage{xcolor}
\usepackage{bm}
\usepackage{amsmath}
\usepackage{textcomp}
\usepackage{siunitx}
\usepackage{graphicx}
\usepackage{caption}
\usepackage{booktabs}
\usepackage{multirow}
\usepackage{tabularx}
\usepackage{subcaption}
\usepackage{tikz} 
\usepackage[ruled,vlined]{algorithm2e}
\usepackage{graphicx,color}
\usepackage{multicol}
\usepackage{natbib}
\usepackage{multirow}
\usepackage{subcaption}
\usepackage{algorithmic}
\usepackage{threeparttable}
\usepackage{booktabs,caption}
\usepackage{threeparttable, tablefootnote}
\usepackage[ruled,vlined]{algorithm2e}
\usepackage{etoolbox}

\usepackage{soul}
\usepackage{xcolor}

\ifCLASSINFOpdf
 
\else
 
\fi

\begin{document}

\title{Multi-level Feature Learning on Embedding Layer of Convolutional Autoencoders and Deep Inverse Feature Learning for Image Clustering}

\author{Behzad Ghazanfari, Fatemeh Afghah\\
%\{bg697,fatemeh.afghah\}@nau.edu\\
School of Informatics, Computing, and Cyber Systems, \\Northern Arizona University, Flagstaff, AZ 86001, USA. 
}

\maketitle

\begin{abstract}

This paper introduces Multi-Level feature learning alongside the Embedding layer of Convolutional Autoencoder (CAE-MLE) as a novel approach in deep clustering. We use agglomerative clustering as the multi-level feature learning that provides a hierarchical structure on the latent feature space. It is shown that applying multi-level feature learning considerably improves the basic deep convolutional embedding clustering (DCEC). CAE-MLE considers the clustering loss of agglomerative clustering simultaneously alongside the learning latent feature of CAE. In the following of the previous works in inverse feature learning, we show that the representation of learning of error as a general strategy can be applied on different deep clustering approaches and it leads to promising results. We develop \textit{deep inverse feature learning (deep IFL)} on CAE-MLE as a novel approach that leads to the state-of-the-art results among the same category methods. The experimental results show that the CAE-MLE improves the results of the basic method, DCEC, around 7\% -14\% on two well-known datasets of MNIST and USPS. Also, it is shown that the proposed deep IFL improves the primary results about 9\%-17\%.  Therefore, both proposed approaches of CAE-MLE and deep IFL based on CAE-MLE can lead to notable performance improvement in comparison to the majority of existing techniques.  The proposed approaches while are based on a basic convolutional autoencoder lead to outstanding results even in comparison to variational autoencoders or generative adversarial networks.

\end{abstract}


\IEEEpeerreviewmaketitle

\section{Introduction}

\begin{figure*}
\centering
\includegraphics[width=2\columnwidth,height=6.2cm]{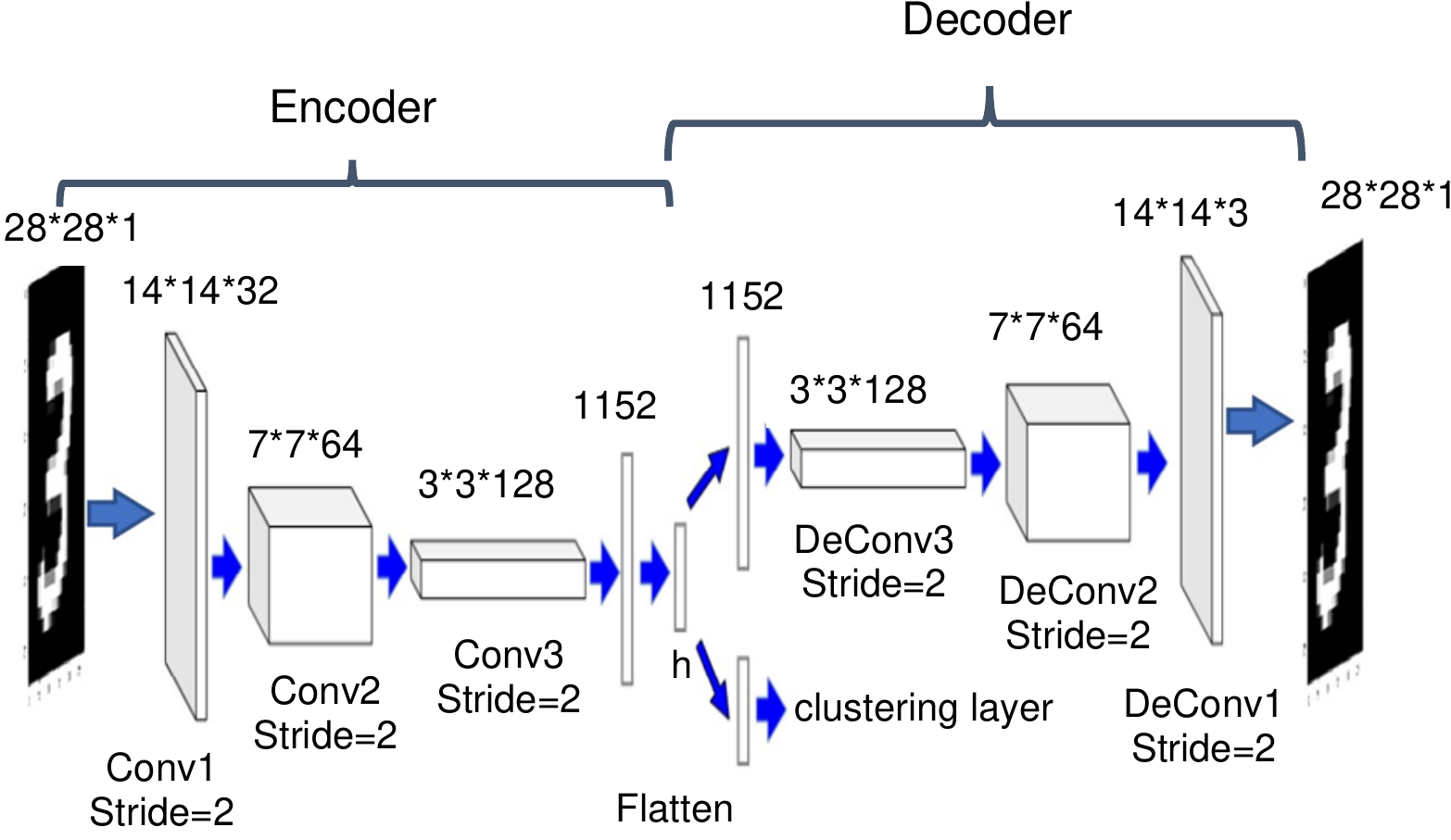}
 \caption{ The network structure of a CAE that proposed in DCEC which learns the latent feature space, $Z$, and provides the clustering layer. The convolutional layers are shown with ``Conv'' and convolutional transpose layers are shown with ``DeConv''. There are fully connected layers in the middle of CAE. We introduce CAE-MLE as the CAE with AC as a multi-level feature learning on the embedding layer.}
\label{fig:CAE}
\vspace{-0.4 cm}
\end{figure*}

While images with labels are limited and expensive, unlabeled images can be obtained much cheaper and on a large scale. It is promising to provide a decision-making mechanism on the huge available unlabeled images or visionary sensory input. Clustering methods as a class of unsupervised learning approaches attempt to divide the instances into several groups based on similarity measures without considering their labels. Clustering methods can be categorized into classical and deep clustering approaches. Deep clustering methods as a sub-category of unsupervised representation learning methods learn high-level representative forms of input data. High-level representative forms are obtained by the layers of deep structures that provide complex non-linear transformations \citep{ghazanfari2020deep}. Deep clustering shows promising performances especially in confront of the datasets with high dimensions like images. Thus, we focus on clustering with deep learning approaches to cluster image datasets in this paper.

Deep clustering methods depending on their architectures or their loss functions can be categorized into several groups \citep{min2018survey}. Two of the most known architectures are based on autoencoders or deep generative approaches. The deep generative approaches include generative adversarial networks (GANs) and variational autoencoders (VAEs). The generative models can produce new samples from the distributions of the obtained clusters. The sample generation can provide a preferred performance in comparison to discriminated ones. Convolutional Neural Networks (ConvNets) as a network structure has a known capability to learn the latent patterns of the images that are used in a variety of architecture for clustering or classification. ConvNets is one of the most known network structure in different deep clustering architectures such as autoencoders, GANs, and VAEs for image datasets. Autoencoder that is based on ConvNets called Convolutional autoencoder (CAE). 

Classical clustering methods are utilized in some deep clustering approaches in different ways. Classical methods apply to the latent space, a joint process of feature learning and clustering, or they are used for pre-training layers \citep{min2018survey}. Most of the approaches which work on the latent feature spaces use k-means as a partitional clustering in which k-means is a flat approach. There are few papers that use spectral or hierarchical clustering in the processing of latent feature space \citep{min2018survey}. Agglomerative clustering (AC) provides interesting abilities to be used jointly alongside the ConvNets for image clustering \citep{yang2016joint}; however, to the best of our knowledge, a multi-level feature learning such as AC  by merging has not been applied on the embedding layer in deep clustering methods. The proposed method in \citep{mautz2019deep} utilizes a tree structure that stands on splitting the latent feature space.

Deep embedded clustering algorithms generally work based on two phases that can be done sequentially or simultaneously. The first phase is learning the latent feature space. The second phase is feature refinement and cluster assignment on the embedded feature space. In this paper, we present and apply a multi-level feature learning based on merging on CAE and the objective function of Deep Convolutional embedded clustering (DCEC) \citep{guo2017deep}, as the basic autoencoder, with some modifications. The DCEC utilizes the CAE structure to learn an embedded feature space in an end-to-end way. The reconstruction loss of the ConvNet and k-means clustering loss are minimized by the stochastic gradient descent and backpropagation. 

We propose the \textbf{CAE-MLE} approach as a multi-level feature learning on the embedding layer of CAE. AC acts as the multi-level feature learning on latent feature spaces of CAE which  its loss function is simultaneously applied alongside the reconstruction loss of CAE. Multi-level feature learning approach is promising for image clustering since the patterns in images generally can be decomposed to several sub-high-level patterns in which some of these sub-high-level patterns are shared among different groups of images. The relations of these sub-high-level patterns or their ratio to each other form a considerable portion of different groups of images. Thus, the hierarchical structure of a multi-level feature learning is more compatible to learn in a down-top manner such inherent structures of the latent features space among different groups of images. We use agglomerative clustering based on ward linkage on the latent feature space as the multi-level feature learning.  To the best of our knowledge, this is the first work in the literature that introduces such a mechanism. Also, we extend deep IFL based on the novel proposed method and introduce a novel feature, \textit{weight of the closest one}, that captures the representation of error of the latent feature space alongside \textit{confidence} and \textit{weight} \citep{ghazanfari2020deep}. We evaluate the contributions on two known image data set MNIST and USPS in which most of approaches have been evaluated on them.     

In summary, we proposed two contributions. First, we propose CAE-MLE  as a multi-level feature learning on the embedding layer of CAE. AC does the multi-level feature learning role. Second, we extend the deep IFL based on the proposed CAE-MLE. The deep IFL as a deterministic solution leads to superior results in comparison to most of the methods in the literature even GANs and VAEs.  In this paper, we introduce deep IFL while we do not use data augmentation and develop this model based on a deterministic approach. Deep IFL in comparison to deterministic models that do not use data augmentation provides the state of the art. Deep IFL also outperforms variational autoencoders or generative adversarial networks that do not use data augmentation except one approach based on a generative adversarial network in one dataset.

\section{Related Works}\label{sec:Related_Works}

There are a variety of methods in deep clustering \citep{min2018survey}. They can be divided into deterministic or generative ones. Generally, the generative approaches like VAEs or GANs lead to better result in comparison to deterministic ones.  Especially, the approaches that are based on GANs lead to the state of the art results in image datasets since they provide and use noise for training too. Also, these clustering approaches can be categorized depending on whether they are using data augmentation or noise in their training or not. Clearly, the approaches that use data augmentation and noise in their training lead to better results. 

Stacked auto-encoders (SAEs) need to be pre-trained layer-per-layer. Then, they are fine-tuned in an end-to-end manner \citep{hinton2006reducing,xie2016unsupervised,guo2017improved}. Stacked denoising auto-encoders (SDAEs) \citep{vincent2010stacked} corrupt the inputs randomly and deliberately by using noise to provide robustness and better results than SAEs. Infinite ensemble clustering (IEC) \citep{liu2018infinite} proposed the marginalized denoising auto-encoder to learn the concatenated deep features and apply k-means on it for their final clustering. 

Joint unsupervised learning (JULE) \citep{yang2016joint} used ConvNets and agglomerative clustering jointly in a recurrent approach. Clustering convolutional neural network (CCNN) \citep{hsu2017cnn} obtains initial cluster centroids. Then, it uses mini-batch k-means that updates the clusters  alongside the parameters of ConvNets by using a stochastic gradient descent approach \citep{min2018survey}. Deep tensor kernel clustering (DTKC) \citep{trosten2020deep} considers persevering consistent cluster structure through ConvNets. 

Deep multi-manifold clustering (DMC) uses ``deep neural network to classify and parameterize unlabeled data which lie on multiple manifolds'' \citep{chen2017unsupervised}.  

Deep embedding clustering (DEC) \citep{xie2016unsupervised} learns the latent feature space with stacked autoencoder and uses k-means clustering with KL divergence based on two steps that are repeated alternately. Improved DEC (IDEC) \citep{guo2017improved} extends DEC by involving the reconstruction loss alongside the clustering loss in the objective function. Discriminatively boosted image clustering with fully convolutional auto-encoders (DBC) \citep{li2018discriminatively} and DCEC \citep{guo2017deep} made extensions of DEC and IDEC by replacing SAEs with CAE respectively in an end-to-end way. In fact, DBC like DEC considers clustering loss, and DCEC considers the reconstruction loss and clustering loss jointly like IDEC. 

Deep embedded regularized clustering (DEPICT) \citep{ghasedi2017deep} is based on CAE with specific characteristics. First, a multinomial logistic regression function stacked on top of CAE. Second, regularized functions are used in the clustering objective function and reconstruction loss. Third, a noisy encoder is used to improve robustness. The proposed method in \citep{shaol2018deep} is an extension of DEPICT that uses a discriminative loss function to enlarge the inter-cluster variation and using the auxiliary data distribution function of DEC to minimize the intra-cluster variation. 

Deep convolutional embedded clustering algorithm with inception like block (DCECI) was introduced in \citep{wang2018deep} in which ``inception-like block with different types of convolution filters'' is designed for local structure preservation of ConvNets. Deep k-means \citep{fard2018deep} reparametrizes the objective function continuously to cluster and learn latent feature space jointly in a better way. Deep autoencoder mixture clustering (DAMIC) \citep{chazan2019deep} associates each cluster with an autoencoder. The clustering tries to learn the nonlinear data representation and the set of the autoencoder. Deep stacked sparse embedded clustering (DSSEC) \citep{cai2019stacked} attempts to preserve the local structure of ConvNets and sparse characteristics of input data.  

Deep adaptive image clustering (DAC) \citep{chang2017deep} works based on a binary pairwise-classification problem in which ConvNets are used to learn label features of images and then convert them to one-hot vectors while it uses data augmentation. Deep continuous clustering (DCC) \citep{shah2018deep} is a CAE that stands on robust continuous clustering that considers clustering as a continuous optimization. The optimization function is based on reconstruction loss, data loss, and pairwise loss. The authors in \citep{tzoreff2018deep} optimize an auto-encoder by using a discriminative pairwise loss function during the auto-encoder pre-training phase. Clustering-driven deep embedding with pairwise constraints (CPAC) \citep{fogel2019clustering} is based on a Siamese network.

SpectralNet \citep{shaham2018spectralnet} is based on the spectral clustering and a Siamese network and uses k-means in the last step. Spectral clustering via ensemble deep autoencoder
learning (SC-EDAE) \citep{affeldt2019spectral} combines ``the spectral clustering and deep autoencoder strengths in an ensemble learning framework''. The authors in \citep{ren2020deep} proposed a two-stage deep density-based image clustering (DDC). DDC works based on CAE in the first stage and then applies t-SNE on the latent feature space. The second stage utilizes a density-based clustering. 

Structural deep clustering network (SDCN) \citep{bo2020structural} proposes some operators to transfer the learned representation by the autoencoder to graph convolutional network. Deep weighted k-subspace clustering (DWSC)  \citep{huang2019deep} is based on an autoencoder and weighted k-subspace network. Concrete k-means (CKM) \citep{gao2020deep} have developed a gradient-estimator for the non-differentiable k-means objective by the Gumbel-Softmax reparameterization way. 

Information maximizing self-augmented training
(IMSAT) \citep{hu2017learning} works based on a fully connected network and regularized information maximization to learn a probabilistic classifier. The classifier tries to maximize the mutual information between the inputs and cluster assignments. The most effective elements in the performance return to a data augmentation known as self-augmented training (SAT). Deep embedded clustering with data augmentation (DEC-DA) \citep{guo2018deep} enhances deep clustering with data augmentation. Data augmentation is used in supervised deep learning approaches to improve the generalization for image datasets. Deep embedded cluster tree (DeepECT) \citep{mautz2019deep} is the first paper that utilizes a hierarchical structure by a cluster tree rather than a flat clustering strategy. DeepECT works based on a hierarchical clustering on embedding that stands on splitting while we believe the latent feature space is learned in CAE should be processed in a multi-level feature learning by merging. The optional data augmentation in DeepECT improves performance --- DeepECT+DA. These papers show the strength of data augmentation, but data augmentation is limited mostly to image datasets.

Two of the known generative architecture in deep learning are GANs and VAEs. VAE models' regularized the encoding distribution during training that helps the latest feature space to provide a proper generative model. Variational deep embedding (VaDE)  \citep{jiang2016variational} and Gaussian Mixture VAE (GMVAE) \citep{dilokthanakul2016deep} are based on VAEs. Latent tree variational autoencoder (LTVAE) \citep{li2018learning} assumes that the latent variable follows a tree structure model. It iterates to improve its learning of the structure of data and capture the facets of data. The proposed method in \citep{yang2019deepDGG} is based on a Gaussian mixture VAE with graph embedding (DGG). In fact, their approach unifies model-based and similarity-based for clustering. The authors in \citep{cao2020simple} proposed a simple, scalable, and stable deep clustering based on a variational deep clustering. Variational autoencoder with distance (VAED) \citep{lim2020deep} extends the work in \citep{song2013auto} by using KL divergence to provide the probability distributions and using Bayesian Gaussian mixture model to cluster the latent feature space.

GANs models constituted of two networks that one of them is generative. The two networks compete and learn from their competition. Adversarial autoencoder (AAE) \citep{makhzani2015adversarial}, deep adversarial clustering (DAC) \citep{harchaoui2017deep}, categorical generative adversarial network
(CatGAN) \citep{springenberg2015unsupervised}, and information maximizing generative adversarial network (InfoGAN) \citep{chen2016infogan} are based on GANs. The authors in \citep{yang2019deep} proposed a ``joint learning framework for discriminative embedding and spectral clustering'' by using a dual CAE that enhanced with improved reconstruction loss and mutual information. Adversarial deep embedded clustering (ADEC) \citep{zhou2019deep} preserves the data structure with AAE by coordinating the distribution of feature representations with the given prior distribution in which k-means works based on distribution distance metrics.

\section{The proposed methods}

The first contribution of this paper is CAE-MLE as a novel deep clustering with CAE that uses agglomerative clustering based on ward linkage instead of k-means on the embedding layer. In other words, we use CAE of DCEC instead of a fully connected autoencoder of DEC. Also, we consider the reconstruction loss and the clustering loss in the objective function like DCEC that makes clustering have direct effects through learning the latent feature space. Finally, we extend the deep IFL based on the proposed CAE-MLE and we introduce a novel feature of the representation of the error as the second part of contributions.

\subsection{Convolutional Autoencoders}

We followed the CAE and the loss functions of DCEC \citep{guo2017deep} that are briefly explained in the following.  
Autoencoder is a known unsupervised learning approach which is a type of neural network to learn the latent feature space of data based on a reconstruction loss. Autoencoder is constituted of an encoder and a decoder. The encoder is a function $z=f_{\phi}(x)$ that maps the input $x$ into a latent representation $z$ and the decoder is a function that reconstructs the latent representation to input $x'=g_{\theta}(z)$. Parameters $\phi$ and $\theta$ denote the weights of the encoder and decoder networks, respectively. They can be learned generally through the minimizing reconstruction loss in autoencoders. We used the mean square error as the distance measure  \citep{xie2016unsupervised}; thus, the optimization objective is defined as follow:
\[\displaystyle\min_{\phi, \theta}L_{rec}  = min\frac{1}{n} \displaystyle\sum_{i=1}^{n} ||x_i -g_{\theta}(f_{\phi}(x_{i}))||_{2}^{2} .\]

ConvNets is one of the most successful architectures in deep learning, especially in computer vision domains. ConvNets learns the high-level features of the images through the series of layers. Convolution layer provides the characteristics including the weight sharing that reduces the number of weight parameters by sharing the weights of depth neurons, local connectivity of the receptive fields which are capable of learning the relation among neighboring pixels, and handling shift-invariance in images \citep{guo2016deep}. A CAE is defined based on applying the Convolutional functions in the above reconstruction loss formula \citep{guo2017deep}.  

\[f_{\phi}(x) = \sigma (x \ast \phi) \equiv h\]

\[g_{\theta}(h) = \sigma (\theta \ast h)\]
where $x$ and $h$ are the matrices, and $\ast$ is the convolution operator.

DCEC replaces SAEs with CAE. DCEC also optimizes both of the reconstruction loss, $L_r$, and clustering loss, $L_c$, simultaneously. A coefficient $\gamma > 0$ controls the degree of distorting embedded space, $L_r$, in the combination with $L_c$. $L_r$ is calculated based on the input space, $x$, and the reconstructed of that through CAE, $x'$.  The embedded layer in CAE is connected to a clustering layer which performs a soft assignment --- it assigns each latent feature space of $z_i$ of the corresponding input space $x_i$ to a soft label $q_i$ by Student's t-distribution \citep{maaten2008visualizing}. The cluster centers $\{\mu\}_{1}^{s}$ are trainable weights to map $x_i$ to soft labels $q_i$ in which $s$ shows the number of clusters.

\[ q_{ij} = \frac{(1+||z_{i} - \mu_{j}||^{2})^{-1}}{\sum_{i}(1+||z_i-\mu_j||^2)^{-1}}\]
where $q_{ij}$ is the probability that $z_{i}$ belongs to  cluster $j$ \citep{guo2017deep}.
$L_c$ is defined similar to DEC and DCEC \citep{guo2017deep,xie2016unsupervised} in the form of Kullback-Leibler (KL) divergence as explained in the following. To this end, the reconstruction loss of autoencoder is added to the objective and optimized along with the clustering loss simultaneously. Thus, the autoencoder preserves the local structure of data and avoids the corruption of feature space \citep{guo2017deep}.

\[ L = L_r+ \gamma L_c\]
\[ L_r = |x-x'|^2_2\]
\[ L_c = KL (P || Q) =\sum_{i} \sum_{j} p_{ij} log \frac{p_{ij}}{q_{ij}}\]

We used the defined target distribution of \citep{guo2017deep}:

\[ p_{ij} = \frac{q_{ij}^2/\sum_{i}q_{ij}}{\sum_{j} (q_{ij}^{2}\sum_{i}q_{ij})}\]

DCEC unlike DEC does not detach the decoder in the refinement of the features and cluster assignments. We followed optimization DCEC in the update of the autoencoder weights and cluster centers and also updating target distribution.

\subsection{Multi-level Feature Learning on Embedding Layer of CAE (CAE-MLE)}
Classical clustering methods can be classified to different groups based on their underlying clustering approach. For instance, k-means is a  partition-based clustering method, spectral clustering is a density-based approach, and agglomerative clustering is a hierarchical clustering method. Despite the good performance of spectral clustering methods, they cannot be easily scalable to large datasets \citep{xie2016unsupervised}.

Hierarchical clustering provides a nested clustering through merging or splitting clusters consecutively. Agglomerative clustering methods merge two clusters with the largest affinity based on the similarity measures iteratively till reaching a stop criterion --- a bottom-up way of hierarchical clustering. Agglomerative clustering finds two clusters $C_{a}$ and $C_{b}$ based on an affinity function, $A$, which measures the similarity of two clusters by affinity measures. There are several affinity measures based on some linkage criteria including single linkage, complete linkage, average linkage, and ward linkage. Ward's linkage \citep{ward1963hierarchical} considers all clusters to minimize the increase of the sum of squared distance within the clusters ---  minimizing the variance of the clusters that being merged which makes it similar to an agglomerative version of k-means objective function. The ward linkage is calculated as follows:

\[\Delta^{(a,b)}= \frac{n_{a}n_{b}}{n_{a}+n_{b}}\parallel (\mu_{a} , \mu_{b}) \parallel^2\]
where $\mu_{a}$ and $\mu_{b}$ are the centroids of two clusters $c_a$ and $c_b$, and $n_{a}$ and $n_{b}$ are the number of instances of the clusters.

\begin{figure}
\centering
\includegraphics[width=0.9\columnwidth,height=4.2cm]{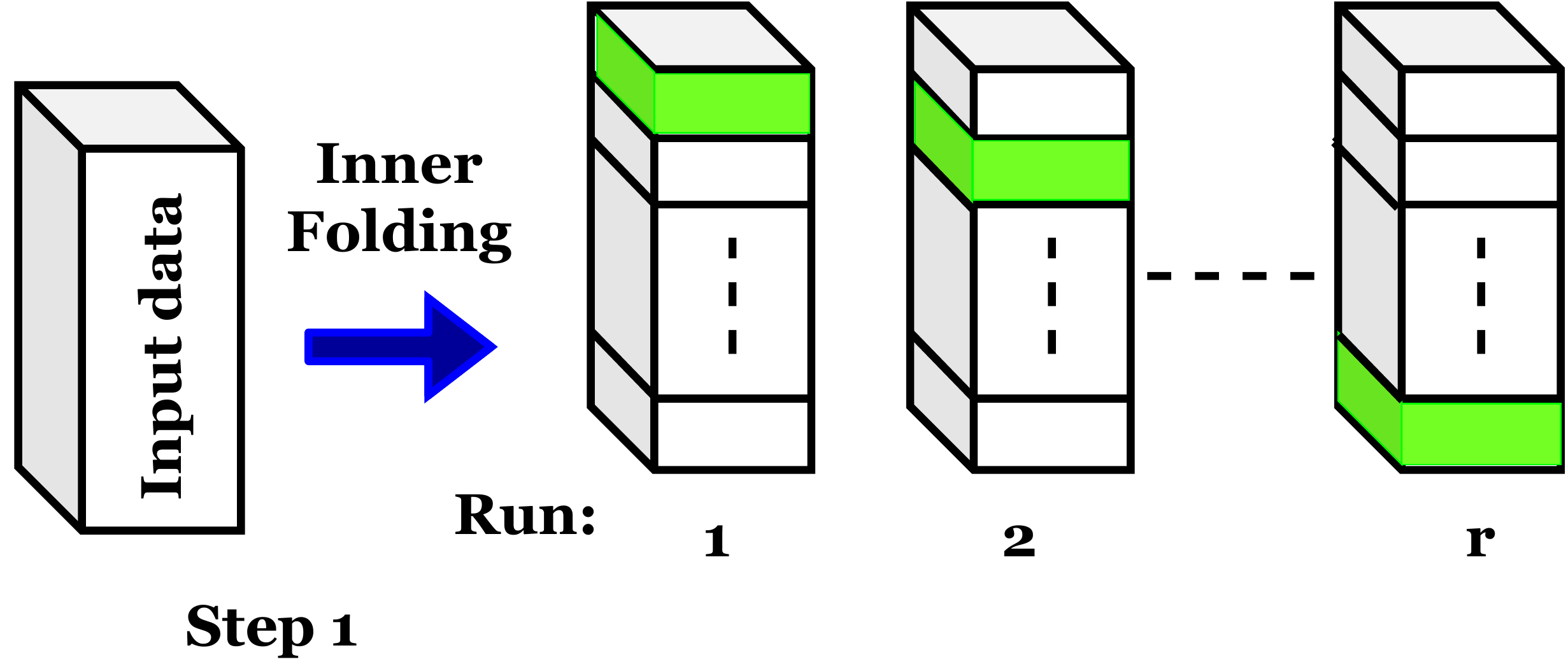}
    \caption{Step 1, inner folding, the input data is partitioned to $r$ non-overlapping folds or partitions. Then, a loop with $r$ runs is applied, where in each run, one fold is considered as an inner-test and $r-1$ remaining folds are considered as inner-train. The inner-test fold in each run is colored with green.}
    \label{fig:Inner_folding_process}
    %\vspace{-0.6 cm}
\end{figure}

\begin{figure*}
\centering
\includegraphics[width=1.9\columnwidth,height=7cm]{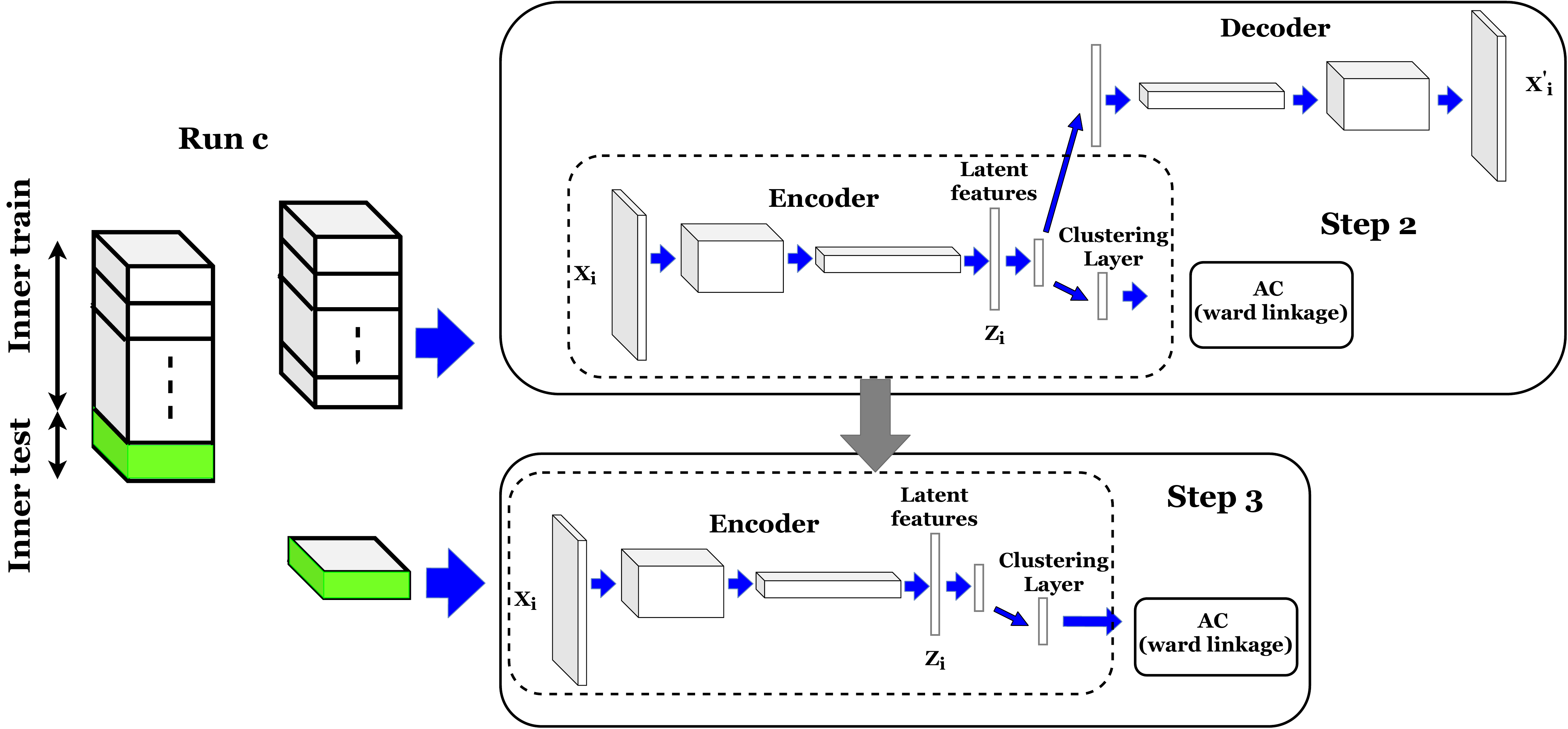}
    \caption{Steps 2 and 3: \textit{Learning the clustered representation of the inner-train data} as step 2 by using the CAE-MLE to train the autoencoder and cluster the latent feature space and obtain their centers. The trained encoder is used in step 3 to  \textit{extract the latent features of inner-test} and the centroids of inner-test as step 3.}
    \label{fig:diagram_23}
    \vspace{-0.1 cm}
\end{figure*}

The hierarchical structure provided by agglomerating clustering with  ward linkage is more compatible with the image datasets. Image datasets are constituted of the significant sub-patterns which are shared among different groups. CAE learns the latent feature space of the images in which the latent feature spaces are the significant sub-patterns of the images that are shared among different groups. Now, AC based on ward linkage is applied  on the latent feature space that is more capable to cluster based on the combination of the shared sub-patterns, the ratio, and the relations among sub-patterns.  Thus, the latent feature space captures more distinguished patterns and the resultant clustering is more powerful.

When CAE is combined with agglomerative clustering as a multi-level feature learning to learn the latent feature space simultaneously with clustering, the embedded features can capture more sophisticated sub-patterns. The reason is that the agglomerative clustering can distinguish more different and sophisticated groups based on a combination of the main latent feature spaces that provides a space for CAE to learn more key patterns in the latent feature space. 

Several groups can be built based on some of these sub-patterns through levels like 2 and 3 that share the convex or 1 and 4 that share the main vertical line and the smaller oblique line, or 6 and 9 that share the circle.  Thus, agglomerative clustering is used in the deep clustering that leads the embedding to learn more key and distinguished patterns in latent feature spaces. Agglomerative clustering based on ward linkage is compatible with this fact that the hierarchical structure can help to learn the sophisticated patterns that some sub-patterns among them are common. In fact, the structure of multi-level is involved in learning to capture the relation among sub-patterns and different groups. 

\subsection{Deep Inverse Feature Learning based on CAE-MLE}

Deep IFL \citep{ghazanfari2020deep} introduces new perspectives about the error and developed in a general form by a basic deep clustering that is based on a fully-connected autoencoder. First, in this paper, we proposed CAE-MLE as a deep clustering approach for image clustering. Then, we extend the previous work deep IFL by replacing CAE-MLE instead of the fully-connected autoencoder and introducing a new feature that can capture the representation of error.

The error in most machine learning methods including both supervised and unsupervised approaches is calculated in the same traditional way of the difference between the predicted and true labels in supervised learning and as the clustering losses in clustering ones. While the representation learning of error introduces that error has other characteristics. The characteristics describe how instances are placed in the feature space in relation to other instances of different clusters which is measured by the level of their similarities and the degree of accuracy of assumptions. In other words, this new perspective considers the error as a dynamic quantity depending on other instances and clusters. IFL methods \citep{ghazanfari2020inverse,ghazanfari2020deep} generates the error and captures the relationships among the instances and the predicted clusters in clustering. Then, they transform the error as high-level features to represent the characteristics of each instance related to the available clusters. The details of the novel perspective of error in deep inverse feature learning in comparison to the error in the literature of machine learning are discussed in \citep{ghazanfari2020deep}.

In this work, we extend our previously proposed deep IFL \citep{ghazanfari2020deep} based on the proposed CAE-MLE for the image datasets. CAE can learn a latent feature space which is much smaller than the original feature set. CAE and agglomerative clustering lead to a better deep embedding clustering since agglomerative clustering handles some portion of the pattern learning in its own structure. In other words, CAE does a nonlinear mapping of the original features to a latent feature space, $Z$, and agglomerative clustering clusters the latent space to $s$ clusters by using cluster assignment hardening to generate centroids, $\mu$, of the clusters. The network structure of CAE is shown in Figure \ref{fig:CAE}.

The deep IFL is described  in \citep{ghazanfari2020deep} based on four main steps that we follow them based on CAE-MLE instead of DEC with some modifications. We briefly explain them in the following  --- for more details and formal definitions, please refer to \citep{ghazanfari2020deep}. 

\noindent \textbf{Step 1- Inner folding}: Inner folding performs a trial process to provide similar conditions to capture error characteristics for all instances. Inner folding is similar to $k$-fold cross-validation, where the data set is partitioned to $r$ non-overlapping folds. In each run, one fold of data is considered as \textit{inner-test} and other folds as \textit{inner-train}. The process is shown in Figure \ref{fig:Inner_folding_process}. The objective of the inner folding process is to obtain the representation of inner-test instances based on the clustered representation of inner-train data.

\noindent \textbf{Step 2- Learning the clustered representation of the inner-train}: The \textit{inner-train} data is fed as input to the CAE-MLE in each round of the inner-training process. CAE-MLE learns the latent feature space, $Z_{train}$, and clustering the inner-train instances with agglomerative clustering in which the objective considers both of the reconstruction loss, $L_r$, and the clustering loss, $L_c$. The encoder of the CAE-MLE is used in step 3 and the clusters and their centroids are utilized in step 4 of the IFL. Step 2 and 3 are depicted in Figure \ref{fig:diagram_23}.

\noindent \textbf{Step 3- Extracting the latent features of inner-test:} The inner-test instances are given as input to the decoder of the CAE-MLE that is trained with inner-train instances to obtain the latent features of inner-test and the clustered representation of them. In other words, the encoder part of autoencoder which was trained in the first phase of CAE-MLE, $f_{\phi}$, is fed with the inner-test instances of the run, $X_{inner\ test}^{J}$, to obtain the corresponding latent features of inner-test instances, $Z^{J}_{inner\_test}$, and the centroids of the clusters, $\{\mu\}^s_1$, as shown in step 3 in Figure \ref{fig:diagram_23}. 
 
\noindent\textbf{Step 4- Feature learning:} In this step, the clusters' centroids of the inner-train which were calculated in step 2 and the latent features of inner-test which were obtained in step 3 are used here to measure several features for each inner-test instance. The relations between the latent features of the inner-test and the clusters' centroids of the inner-train are calculated and considered as new features for that inner-test sample. Thus, we extract a new set of features for each inner-test instance during each run of inner folding process.

\begin{algorithm}
\caption{Deep Inverse feature learning based on CAE-MLE for clustering.}\label{alg:IFL_cluster}
\begin{algorithmic}[1]

\BlankLine

\STATE \textbf{Input}$:$ $X$.
\STATE \textbf{Output}$: $ \textit{Features of error representation for $X$}.

\STATE{ \textbf{Step 1: Inner folding}} % 
// Partition $X$ to $r$ folds and perform $r$ runs, where in each round, $r-1$ folds are set as the inner-training samples and the remaining one fold is set as inner-test instances.   

\FOR{$j=1:r$}
        \STATE{inner-test = $J^{th}$ fold of $X$ --- $X^{J}_{Inner\ test}$.}
        \STATE{inner-train = All folds except $J^{th}$ fold of $X$ --- $X^{J}_{Inner\ train}$.}
        \STATE{\textbf{Step 2: Learning the clustered representation of the inner-train}} 
         \STATE{${}$\hspace{0.1cm} \textbf{Input of step 2}$:$ inner-train, $X^{J}_{Inner\ train}$.}
         \STATE {${}$\hspace{0.1cm} \textbf{2.1)} Applying CAE-MLE on inner-train, $X^{J}_{Inner\ train}$.}
         \STATE {${}$\hspace{0.1cm} \textbf{Outputs of step 2}$:$ Encoder of CAE-MLE, $f_{\phi}$, clusters, $\{c_i\}_{i=1}^{s}$, and centroid of clusters --- $\{\mu_i\}_{i=1}^{s}$.}\\
         \STATE{\textbf{Step 3: Extracting the latent features of inner-test}} 
         \STATE{${}$\hspace{0.1cm} \textbf{Inputs of step 3}$:$ Encoder of CAE-MLE, $f_{\phi}$, obtained in step 2, and inner-test --- $X^{J}_{Inner\ test}$.}
         \STATE {${}$\hspace{0.1cm} \textbf{3.1)} Feeding encoder with inner-test, $X^{J}_{Inner\ test}$.}
         \STATE {${}$\hspace{0.1cm} \textbf{Output of step 3}$:$ Latent features of inner-test --- $Z^{J}_{Inner\ test}$, and their corresponding centroid of clusters --- $\{\mu_i\}_{i=1}^{s}$.}
         
         \STATE {\textbf{Step 4: Feature learning}}  
         \STATE{${}$\hspace{0.1cm} \textbf{Inputs of step 4}$: $  Clusters, $\{c_i\}_{i=1}^{s}$, and their centroid, $\{\mu_i\}_{i=1}^{s}$, and latent features of inner-test, $Z^{J}_{Inner\ test}$}.
          \STATE {${}$\hspace{0.1cm} \textbf{4.1)} Calculating ``confidence'', ``weight'', and ``weight of the closest one'' for each inner-test instance.}
         \STATE{${}$\hspace{0.1cm} \textbf{Output of step 4}$: $ Features of error representation for the inner-test or $J^{th}$ fold of $X$ --- $X^{J}_{Inner\ test}$}.
 \ENDFOR     
\end{algorithmic}
\end{algorithm}

In other words, the representation of one set, inner-test instances or test instances, is evaluated based on the clustered representation of other set instances, inner-train,  or training. We assign each inner-test instance to all possible clusters and measure the representation of such assignment on the clustered representation and map that in the form of features as the corresponding error representation of that inner-test instance for that cluster. Here, we introduce three metric, \textit{confidence},  \textit{weight}, and \textit{weight of the closest one} which are measured per instance of each cluster, $\{c_i\}_{i=1}^s$. 

\textit{\textbf{Confidence}} is the ratio of the number of elements in a  cluster in which the inner-test instance has the closest distance to its center to the  number of all instances in all clusters. Thus, we need to find the closest cluster by measuring the distance between the latent space of the instance to the centroids of the clusters which are obtained from the CAE-MLE and calculating the number of instances which belongs to that cluster to the  number of all instances. If $dist(z_j, c_b)= min (dist(z_j, \{c_i\}_{i=1}^{s}))$,  $confidence(x_{j})=\frac{|c_{b}|}{|C|}$ in which $z_j$ is the corresponding latent feature of $x_j$. Confidence is one feature. 

\textit{\textbf{Weight}} of belonging an instance in inner-test to a cluster on inner-train is the distance of the latent feature of the element to the center of the cluster, $\mu$. We calculate the \textit{weight} for each inner-test and the centers of all clusters, $\{\mu\}_{i=1}^{s}$, of inner-train. In other words, \textit{weight} is a vector in length of number of clusters in which $weight(x_{j}, \{\mu\}_{i=1}^{s})$= dist $( z_{j},\{\mu\}_{i=1}^{s})$. Thus, the number of features of weight is the same as the number of clusters. 

\textit{\textbf{Weight of the closest one}} is the distance of the latent feature of each element of inner-test to a center of the cluster on inner-train instances, $\mu$, that has the lowest distance to it. We calculate the \textit{Weight of the closest one} for each inner-test and the closest center of all clusters, $\{\mu\}_{i=1}^{s}$, of inner-train. In other words, \textit{weight of the closest one} is a number in which $weight-closest(x_{j}, \{\mu\}_{i=1}^{s})$= $min_i (dist ( z_{j},\{\mu\}_{i=1}^{s})$. Weight of the closest one is one feature. 

\begin{table}[t]
%\vspace{-10pt}
\caption{ Descriptions of the data sets.}
\label{datasetsdesriptoin}
%\vskip 0.15in
\begin{center}
\begin{small}
\begin{sc}
\resizebox{1\columnwidth}{!}{
\begin{tabular}{lrcr}%|cccr
\toprule
\textbf{Data sets} & \textbf{\#instances} & \textbf{\#features} & \textbf{\#classes}\\ %& Data sets & \#inst. & \#feat. & \#class \\
\midrule
\hline
MNIST    & 70000 & 26*26 (784) & 10 \\
\hline
USPS    & 9298 &  16*16 (256)& 10 \\
\bottomrule
\end{tabular}}
\end{sc}
\end{small}
\end{center}
%\vskip -0.1in
\vspace{-12pt}
\end{table}

\begin{table*}[!htbp]
\caption{The ACC and NMI of a variety of clustering approaches are shown on MNIST.}
\label{table:MNIST_result_list_1}
\begin{threeparttable}
\vskip 0.15in
%\vskip 0.15in
\begin{center}
\begin{small}
%\begin{sc}
\resizebox{0.75\textwidth}{!}{
\begin{tabular}{|l|cc||l|cc|}%|cccr
\hline
 & \multicolumn{2}{c||}{ \textbf{MNIST}} & & \multicolumn{2}{c|}{  \textbf{MNIST}} \\

\hline
 \textbf{Clustering Models} &  \textbf{NMI} &  \textbf{ACC} &  \textbf{Clustering Models}  &  \textbf{NMI} &  \textbf{ACC}\\
\hline
\hline
K- \MakeLowercase{means}\tnote{1} \; \citep{wang2014optimized} &  49.97 & 57.23 & JULE \; \citep{yang2016joint}    & 91.30  &  96.40  \\\hline
SC\tnote{1}\; \citep{zelnik2005self} &  66.26  & 69.58 & SpectralNet \citep{shaham2018spectralnet} \;   &92.4 &    97.1 \\\hline
AC\tnote{1}\; \citep{gowda1978agglomerative} & 60.94   &  69.53 & DBC \citep{li2018discriminatively}  & 91.7 &   96.4   \\\hline 
AE\tnote{1}\; \citep{bengio2007greedy}    & 72.57   & 81.23 & DDC \citep{ren2020deep} & 93.2 & 96.5   \\\hline
SAE\tnote{1}\; \citep{ng2011sparse}    & 75.65  & 82.71 & FcDEC \citep{guo2018deep} & 87.5 & 91.6 \\\hline
DAE\tnote{1}\; \citep{vincent2010stacked}   & 75.63   & 83.16 & ConvDEC \citep{guo2018deep} & 88.8 & 90.0 \\\hline
DeCNN\tnote{1}\; \citep{zeiler2010deconvolutional}    & 75.77 & 81.79  & FcIDEC \citep{guo2018deep} & 87.2 & 91.2 \\\hline
SWWAE\tnote{1}\; \citep{zhao2015stacked}   & 73.60   & 82.51  & ConvIDEC \citep{guo2018deep} & 89.1 & 90.1\\\hline
AEVB\tnote{1}\; \citep{kingma2013auto}   & 73.64 &  83.17  & FcDCN \citep{guo2018deep} & 84.9 & 90.1  \\\hline
GMM\tnote{2}\;  & -- & 53.73  & Method \citep{tzoreff2018deep} &  --   & 97.4  \\\hline
AE+GMM\tnote{2}\;  & -- & 82.18 & SC-EDAE \citep{affeldt2019spectral} &  87.93  & 93.23 \\\hline
SEC\tnote{3}\; \citep{nie2011spectral}   & 77.9  &  80.4  & S3VDC \citep{cao2020simple} &  --  &  93.60 \\\hline
LDMGI\tnote{3}\; \citep{yang2010image}   &80.2  &  84.2  & DCECI \citep{wang2018deep} &  92.84   & 96.81 \\\hline
DEC\tnote{1}\; \citep{xie2016unsupervised}    & 77.16 &  84.30 &  VAE \citep{kingma2013auto}+GMM \tnote{2} & --  &  72.94 \\\hline
IDEC\; \citep{guo2017improved}    &86.72 &    88.06  &  VaDE \citep{jiang2017variational}  & --   & 94.46 \\\hline
SAE+k-means\tnote{4}  &79.27 & 84.90 &  LTVAE \citep{li2018learning} & --   & 86.32 \\\hline
DCEC \citep{guo2017deep}  & 88.49 & 88.97 &  VAED \citep{lim2020deep}  & 81.9  & 88.75\\\hline
DC-Kmeans \citep{tian2017deepcluster}  &74.48 &  80.15 &  GAN \tnote{1} \; \citep{radford2015unsupervised}  & 76.37  &82.79\\\hline
DC-GMM \citep{tian2017deepcluster}  & 83.18 &  85.55 & AAE \tnote{2}\; \citep{makhzani2015adversarial} & -- &  83.48  \\\hline
Method \citep{gultepe2018improving}  & 82.4  &  88.2 &  GMVAE \citep{dilokthanakul2016deep}  & --  & 82.31 \\\hline
Autoencoder+ k-means\tnote{5}  & --  & 81.84  & Method \citep{shaol2018deep} &  91.9  & 96.9\\\hline
Autoencoder+ LDMGI\tnote{5}  & --  & 83.98 &InfoGAN \citep{chen2016infogan}  &  84.00  &  87.00 \\\hline
Autoencoder+ SEC\tnote{5}  & -- & 81.56 & ClusterGAN \citep{mukherjee2019clustergan} & 89.00  & 95.00 \\\hline
DCN \citep{yang2017towards}  & 81.00 & 83.00 &Conv-CatGAN \citep{springenberg2015unsupervised} & -- & 95.73  \\\hline
IMSAT(RPT) \citep{hu2017learning}  & -- &  89.6  & ADEC \citep{zhou2019deep} & 94.3 &  96.5\\\hline
DSSEC \citep{cai2019stacked} &  85.7  &  87.7 & Conve-ADEC \citep{zhou2019deep} & 95.5 & 97.2 \\\hline
k-DAE \citep{opochinsky2020k} &  86.00   & 88.00  &
DCAE \citep{alqahtani2018deep} & 84.97  & 92.14  \\\hline
DWSC \citep{huang2019deep} & 88.9 &94.8  &
Method \citep{aljalbout2018clustering}  &92.3  & 96.1  \\\hline
IEC \citep{liu2018infinite} & 54.20 & 60.86 & \textbf{Deep IFL (DEC) \citep{ghazanfari2020deep}}  & \textbf{--} & \textbf{95.79}\\\hline
DAMIC \citep{chazan2019deep} & 87.00  & 89.00 & \textbf{CAE-MLE}  & \textbf{92.57} & \textbf{96.77} \\\hline
CKM \citep{gao2020deep} & 81.4  & 85.4  &\textbf{Deep IFL (CAE-MLE)}  & \textbf{93.09} & \textbf{97.50} \\\hline
DeepECT \citep{mautz2019deep} &  --  & 82.00  \\\hline
\bottomrule
\end{tabular}}
%\end{sc}
\end{small}
\end{center}
\begin{tablenotes}\footnotesize
\item [1] Results taken from \citep{chang2017deep}.
\item [2] Results taken from \citep{jiang2017variational}.
\item [3] Results taken from \citep{ghasedi2017deep}.
\item [4] Results taken from \citep{guo2017deep}.
\item [5] Results taken from \citep{xie2016unsupervised}.
\end{tablenotes}
\end{threeparttable}
%\vskip -0.1in
\vspace{-12pt}
\end{table*}

\begin{table*}[!htbp]
\caption{The ACC and NMI of a variety of clustering approaches are shown on USPS. }
\label{table:USPS_result_list_1}
\begin{threeparttable}
\vskip 0.15in
%\vskip 0.15in
\begin{center}
\begin{small}
%\begin{sc}
\resizebox{0.75\textwidth}{!}{
\begin{tabular}{|l|cc||l|cc|}%|cccr
\hline
 & \multicolumn{2}{c||}{ \textbf{USPS}}  & & \multicolumn{2}{c|}{ \textbf{USPS}}\\
\hline
\textbf{Clustering Models} & \textbf{NMI} & \textbf{ACC} & \textbf{Clustering Models} & \textbf{NMI} & \textbf{ACC}\\
\midrule
\hline
k-means\tnote{1} \;  &65.9 &  69.4   & FcDEC \citep{guo2018deep} & 78.4 & 76.3\\\hline
LDMGI \citep{yang2010image}    &86.5 &  81.5  & ConvDEC \citep{guo2018deep} & 82.2 &  78.6\\\hline
GMM \tnote{2} & 21.07&   29.00   &  FcIDEC \citep{guo2018deep} & 79.9 &  77.1\\\hline
DAE+k-means \tnote{2} &52.03&  59.55  &  ConvIDEC \citep{guo2018deep} & 82.7 & 79.2 \\\hline
DAE+GMM \tnote{2} & 59.67 &   64.22  & FcDCN \citep{guo2018deep} & 72.2 &  69.9  \\\hline
DAEC  \tnote{2} \; \citep{song2013auto} &54.49 & 61.11  &  DeepECT \citep{mautz2019deep} &  --  & 72.00  \\\hline
DEC \tnote{2} & 61.91 &   62.46  &  k-DAE \citep{opochinsky2020k} &  80.00  &  77.00   \\\hline
DC-k-means \citep{tian2017deepcluster}  &57.37 & 64.42  & Deep  k-Means \citep{fard2018deep} &  77.6  & 75.70  \\\hline
DC-GMM \citep{tian2017deepcluster}  & 69.39  & 64.76  & DTKC \citep{trosten2020deep} &  73.00 &  70.00 \\\hline
SC-ST\tnote{3} \; \citep{zelnik2005self} &   72.6   & 30.8  & DCECI \citep{wang2018deep} &  79.06  &  81.61 \\\hline
SC-LS\tnote{3} \; \citep{chen2011large} &   68.1    & 65.9  & SC-EDAE \citep{affeldt2019spectral} &  83.17  & 81.78  \\\hline
AC-GDL\tnote{3} \; \citep{zhang2012graph} &  82.4   & 86.7  & VAED \citep{lim2020deep} & 62.33  &  76.13 \\\hline 
AC-PIC\tnote{3} \;\citep{zhang2013agglomerative} & 84.00 &  85.5  & DDC \citep{ren2020deep} & 96.7 & 91.8  \\\hline
SEC\tnote{3} \; \citep{nie2011spectral}    &51.1  & 54.4  &  JULE\tnote{2} \; \citep{yang2016joint}    & 91.3 &  95.0  \\\hline
DEC\tnote{3} \; \citep{xie2016unsupervised}   & 58.6 & 61.9  & Method \citep{alqahtani2019learning}&  -- & 89.23  \\\hline
SAE+k-means\tnote{4}  & 57.27  & 61.65  &  DDC \citep{ren2020deep} & 96.7 & 91.8  \\\hline
CAE+k-means\tnote{4}  & 57.27  &  61.65  & Method \citep{gultepe2018improving}  & 86.8   & 92.6  \\\hline
DCEC \citep{guo2017deep}  &82.57 &  79.00  & ADEC \citep{zhou2019deep} & 80.6 & 81.4 \\\hline
IDEC \citep{guo2017improved} & 78.64 &  76.05  &  Conve-ADEC \citep{zhou2019deep} & 92.7 & 97.1 \\\hline
CKM \citep{gao2020deep} & 70.7 & 72.1  &  NSC-AGA \citep{ji2019nonlinear} & 77.27   & 72.56    \\\hline
AGDL \citep{zhang2012graph} \;   & -- & 82.4  &  \textbf{Deep IFL (DEC)}  & \textbf{--} & \textbf{84.55}  \\\hline
DBC \citep{li2018discriminatively}  & 72.4  & 74.3  & \textbf{CAE-MLE}  & \textbf{87.92} & \textbf{93.53} \\ \hline
CPAC-VGG \citep{fogel2019clustering} & 92.00  & 87.00  &  \textbf{Deep IFL (CAE-MLE)}  & \textbf{92.26} & \textbf{96.89} \\\hline
\bottomrule
\end{tabular}}
\end{small}
\end{center}
\begin{tablenotes}\footnotesize
\item [1] Results taken from \citep{gultepe2018improving}.
\item [2] Results taken from \citep{tian2017deepcluster}.
\item [3] Results taken from \citep{ghasedi2017deep}.
\item [4] Results taken from \citep{guo2017deep}.
\end{tablenotes}
\end{threeparttable}
%\vskip -0.1in
%\vspace{-12pt}
\end{table*}

\section{Experimental Results}

In this paper, we propose our approach based on an autoencoder as a discriminative model in a generic and basic manner. The reason that we select a discriminative autoencoder is that we want to focus on showing the performance of the contributions including ``CAE-MLE'' and ``deep IFL based on CAE-MLE'' while they are independent strategies and can be applied on deterministic or generative models. We evaluate the performance of the proposed feature learning model and compared the proposed method with both discriminative and generative ones. It is shown the CAE-MLE and deep IFL based on CAE-MLE lead to promising results while stands on a basic architecture. We use AC based on ward linkage that is similar to k-means objective function while providing a hierarchical mechanism for multi-level feature learning. 

We evaluated the proposed method on two common image datasets that are described in Table \ref{datasetsdesriptoin}. These data sets are two handwritten digit image data sets (USPS \footnote{http://www.cs.nyu.edu/˜roweis/data.html} and MNIST \citep{lecun1998gradient}).  The details of the data sets and the pre-processing process are described in the following:

\textbf{MNIST:} MNIST data set \citep{lecun1998gradient} includes 70000 handwritten digits with size 28*28 pixels, in which the range of the values is between [0, 255]. The value of each pixel is divided by the max values to normalize them to an interval [0, 1]. 

\textbf{USPS:} USPS data set contains 9298 gray-scale handwritten digits with size 16*16 pixels which are normalized to [-1, 1].

\begin{table}[!htbp]
\caption{The ACC and NMI of clustering approaches that the network structure has tuned per dataset or or DA or noise reconstruction is the superiority part of their mechanisms on MNIST.}
\label{table:MNIST_result_list_3}
\begin{threeparttable}
\vskip 0.15in
%\vskip 0.15in
\begin{center}
\begin{small}
%\begin{sc}
\resizebox{0.85\columnwidth}{!}{
\begin{tabular}{|l|cc|}%|cccr
\toprule
& \multicolumn{2}{c|}{ \textbf{MNIST}} \\\hline
\textbf{Clustering Models} & \textbf{NMI} &  \textbf{ACC} \\
\midrule
\hline
DEPICT \citep{ghasedi2017deep}    & 91.7 &  96.5  \\\hline
DAC \citep{chang2017deep}    & 93.51 &   97.75 \\\hline
Method \citep{yang2019deep} & 94.1  &  97.8  \\\hline
DGG \citep{yang2019deepDGG}  & --   & 97.58  \\\hline
IMSAT(VAT) \citep{hu2017learning}  & -- & 98.4  \\\hline
DeepECT-DA  \citep{mautz2019deep} &  --   & 94.00  \\\hline
FcDEC-DA \citep{guo2018deep} & 95.9 & 98.5 \\ \hline
ConvDEC-DA \citep{guo2018deep} & 96.0& 98.5 \\ \hline
FcIDEC-DA \citep{guo2018deep} & 96.2 & 98.6 \\ \hline
ConvIDEC-DA \citep{guo2018deep} & 95.5& 98.3 \\ \hline
FcDCN-DA \citep{guo2018deep} & 96.2 & 98.6 \\ \hline
DDC-DA \citep{ren2020deep} & 94.1  & 96.9  \\\hline
Conve-ADEC-DA \citep{zhou2019deep} & 97.1 &  99.3 \\ \hline

\bottomrule
\end{tabular}}
%\end{sc}
\end{small}
\end{center}
\end{threeparttable}
%\vskip -0.1in
\vspace{+15pt}
\end{table}

\begin{table}[!htbp]
\caption{The ACC and NMI of clustering approaches that the network structure has tuned per dataset or DA or noise reconstruction is the superiority part of their mechanisms on USPS.}
\label{table:USPS_result_list_3}
\begin{threeparttable}
\vskip 0.15in
%\vskip 0.15in
\begin{center}
\begin{small}
%\begin{sc}
\resizebox{0.85\columnwidth}{!}{
\begin{tabular}{|l|cc|}%|cccr
\toprule
 & \multicolumn{2}{c|}{ \textbf{USPS}} \\\hline
\textbf{Clustering Models} & \textbf{NMI} & \textbf{ACC} \\\hline
\midrule
Method \citep{yang2019deep} & 85.7  &  86.9  \\\hline
DEPICT \citep{ghasedi2017deep}    & 92.7 &   96.4  \\\hline
FcDEC-DA \citep{guo2018deep} & 94.5 & 98.0 \\ \hline
ConvDEC-DA \citep{guo2018deep} & 96.2 & 98.7 \\ \hline
FcIDEC-DA \citep{guo2018deep} & 95.4 &  98.4 \\ \hline
ConvIDEC-DA \citep{guo2018deep} & 95.5 & 98.4 \\ \hline
FcDCN-DA \citep{guo2018deep} & 92.9 & 96.9 \\ \hline
DDC-DA \citep{zhou2019deep} & 94.1 &  96.9 \\ \hline
Conve-ADEC-DA \citep{zhou2019deep} & 96.6 & 99.1 \\ \hline
DeepECT-DA \citep{mautz2019deep} &  --   & 89.00  \\\hline
DDC-DA \citep{ren2020deep} & 97.7 & 93.9  \\
\hline
\bottomrule
\end{tabular}}
%\end{sc}
\end{small}
\end{center}
\end{threeparttable}
%\vskip -0.1in
%\vspace{-12pt}
\end{table}

The common unsupervised clustering accuracy, denoted by ACC, is used to evaluate the performance of this model as defined as following:
\begin{equation}
  \textit{ACC}=\max_{m}\frac{\sum_{i=1}^{n} 1\{y_i=m(c_i)\}}{n}
\end{equation}
where $c_i$ is the cluster assignment for instance $i$ determined by the clustering approach, $y_i$ is the \textit{ground-truth} label or the label of $i^{\text{th}}$ instance, and $m$ is the mapping function that ranges over all possible one-to-one mapping between $c_i$ and $y_i$. 
 
Normalized Mutual Information (NMI) \citep{estevez2009normalized} that are defined as following: 
 
\begin{equation}
  \textit{NMI(Y,C)}= \frac{\textit{I(Y,C)}}{\frac{1}{2}[H(Y)+H(C)]}
\end{equation}
where $I$ is the \textit{mutual information} metric and $H$ is the entropy.

The performance of the proposed model is compared with both the classical clustering methods such as k-means [32], spectral clustering, agglomerative clustering for the ward and average linkage criterion  [33],  and Gaussian  mixture  model  (GMMs)  as  well as  deep  clustering learning  approaches  including auto-encoder (AE)   \citep{bengio2007greedy}, denoising auto-encoder \citep{vincent2010stacked}, autoencoder+k-means, local discriminant models and global integration (LDMGI) \citep{yang2010image}, sparse auto-encoder (SAE) \citep{ng2011sparse}, autoencoder  +  LDMGI \citep{xie2016unsupervised}, deconvolutional network (DeCNN) \citep{zeiler2010deconvolutional},  auto-encoding variational bayes (AEVB)   \citep{kingma2013auto}, stacked what-where auto-encoder (SWWAE) \citep{zhao2015stacked}, autoencoder+GMM, autoencoder+SEC  \citep{xie2016unsupervised},  DEC,  IDEC \citep{guo2017improved}, JULE \citep{yang2016joint}, DCEC \citep{guo2017deep}, DBC \citep{li2018discriminatively}, DAE \citep{vincent2010stacked}, DCN \citep{yang2017towards}, DEPICT \citep{ghasedi2017deep}, IEC \citep{liu2018infinite}, DCC \citep{shah2018deep}, CCNN \citep{hsu2017cnn}, SDCN \citep{bo2020structural}, CKM \citep{gao2020deep}, DAMIC \citep{chazan2019deep}, k-DAE \citep{opochinsky2020k}, DWSC \citep{huang2019deep}, DCECI \citep{wang2018deep}, DTKC \citep{trosten2020deep}, S3VDC \citep{cao2020simple}, CPAC-VGG \citep{fogel2019clustering}, FcDEC \citep{guo2018deep} , ConvDEC \citep{guo2018deep}, FcIDEC \citep{guo2018deep}, ConvIDEC \citep{guo2018deep}, FcDCN \citep{guo2018deep}, SC-EDAE \citep{affeldt2019spectral}, DDC \citep{ren2020deep}, and SpectralNet \citep{shaham2018spectralnet}.

The approaches that are based on generative models GANS or VAEs including GAN \citep{radford2015unsupervised}, adversarial autoencoder (AAE) \citep{makhzani2015adversarial}, Conv-CatGAN \citep{springenberg2015unsupervised}, InfoGAN \citep{chen2016infogan}, ClusterGAN \citep{mukherjee2019clustergan}, ADEC \citep{zhou2019deep}, Conve-ADEC \citep{zhou2019deep},  VAE+GMM, GMVAE \citep{dilokthanakul2016deep},  VaDE \citep{jiang2017variational}, LTVAE \citep{li2018learning}, DGG \citep{yang2019deepDGG}, and VAED \citep{lim2020deep}. The approaches that are using data augmentation (DA) as the main part of their models including IMAST (VAT) \citep{hu2017learning}, DAC \citep{chang2017deep}, FcDEC-DA \citep{guo2018deep}, ConvDEC-DA \citep{guo2018deep}, FcIDEC-DA \citep{guo2018deep}, ConvIDEC-DA \citep{guo2018deep}, FcDCN-DA \citep{guo2018deep}, DDC-DA \citep{zhou2019deep}, Conve-ADEC-DA \citep{zhou2019deep}, and DeepECT-DA \citep{mautz2019deep}. 

Here, we compare the performance of the proposed CAE-MLE and deep IFL based on CAE-MLE, deep IFL (CAE-MLE), with the state-of-the-art techniques in clustering based on ACC and NMI. It should be noted that the deep IFL features are highly abstract, (i.e., only $2+s$ in which $s$ shows the number of clusters). In MNIST and USPS, the numbers of primary features are $784$ and $256$, respectively while the number of deep IFL features is $12$ that are added to the primary features. In this paper, we set the number of runs for the inner folding process to 10 (i.e., $r=10$) for all experiments and we repeat the experiments 5 times to report the average of the results. In this paper, we used the autoencoder network that is based on convolution layers and convolution transpose layers in 9 layers that their dimensions are shown in Figure \ref{fig:CAE} --- following DCEC \citep{guo2017deep}. The learned features are normalized between [-2.5, 2.5].

Tables \ref{table:MNIST_result_list_1} and \ref{table:USPS_result_list_1}  show the results of a varieties  of approaches including deterministic models, VAE, or GANs for MNIST and USPS correspondingly. To have a better comparison of the similar approaches, we bring the approaches that DA or noise reconstructions are the superiority parts of them in separated tables --- Tables \ref{table:USPS_result_list_3} and \ref{table:MNIST_result_list_3}. Clearly, there are some approaches based on GANs that use noise and DA or some approaches that are based on DA in the first tables of datasets. In these tables, ``--'' refers to the cases in which the results were not reported. In Tables \ref{table:MNIST_result_list_3} and \ref{table:USPS_result_list_3}, it can be seen that most approaches that are based on DA or noise reconstructions lead to promising results while such techniques limit the generality of their approaches on different datasets.

As shown in Table \ref{table:MNIST_result_list_1}, the idea of using multi-level feature learning on CAE, CAE-MLE, improves NMI and ACC of DCEC around 4\% and 7\% that makes it close to methods that are even based on GAN or VAE. Deep IFL (CAE-MLE) leads to the state of the art among deterministic and generative models including GANs or VAE. In Table \ref{table:MNIST_result_list_3}, it can be seen most approaches that are based on DA or noise reconstructions lead to promising results while such techniques limit the generality of their approaches on different kinds of datasets like text \citep{ghazanfari2020deep}. 

As shown in Table \ref{table:USPS_result_list_1}, CAE-MLE improves NMI and ACC of DCEC around 5\% and 5\% for USPS. It is shown that deep IFL (CAE-MLE) provides more improvement, especially in cases where the dataset is not as large as MNIST. Deep IFL (CAE-MLE) provides superior performance over deterministic and generative models except for Conve-ADEC as a GAN-based model with 2\% differences. It can be justified that Conve-ADEC as a generative model can show the superiority of generation of data especially for small datasets.  

Among the methods mentioned in Table \ref{table:USPS_result_list_3}, DeepECT which uses hierarchical processing on the latent feature space based on splitting is the most similar approach to the proposed CAE-MLE. However, the proposed CAE-MLE leads to considerably better results compared to DeepECT --about 15\% and 21\% for MNIST and USPS, respectively. It justifies the proposed merging point of the latent feature space for different groups in CAEs for image clustering. This improvement is achieved because of the proposed deep IFL as a general representation learning strategy which can be even applied on the basic clustering methods and leads to the state of the art even in comparison to GANs and VAE models.

In summary, we can conclude that the CAE-MLE and deep IFL based on CAE-MLE provide new perspectives that offer considerable accuracy improvement based on the extensive experimental results. We showed the idea of using AC on the CAE instead of k-means that improves the results of DCEC, unsupervised clustering accuracy, around 8\% and 14\% in USPS and MNIST datasets correspondingly. Also, we showed deep IFL based on CAE-MLE provides superior performance than most papers in the literature even better than generative ones. Deep IFL leads to the state of the art among the discriminative approaches and its results are just below one method on one dataset that stands on a GAN. We have been focused on the improvement that AC provides on the embedding layer of the CAE and the representation of error rather than attempting to develop the most optimal network architecture.

\section{Conclusion}

In this paper, we introduce \textit{CAE-MLE} as a novel multi-level feature leaning on the embedding layer of CAE. We use AC as the multi-level feature learning that works simultaneous with learning latent feature space --- $L = \gamma L_c + L_r$. AC based on ward linkage is similar to k-means objective function while providing a hierarchical mechanism for multi-level feature learning.  Image datasets patterns are generally constituted of sub-patterns in which these sub-patterns are shared among different groups; thus, AC's structure can capture such relation to learn precisely different groups for image datasets. In other words, CAE-MLE leads the embedding layer to capture more key and distinguished patterns in the latent feature space. It is shown in the experimental results that the proposed contribution considerably improves the results of the basic method, DCEC, on two known datasets. The strategy of inverse feature learning can be implemented by different learning approaches as tactics \citep{ghazanfari2020inverse,ghazanfari2020deep}. We extend the deep IFL in \citep{ghazanfari2020deep} based on the proposed CAE-MLE and we introduce a novel feature, \textit{weight of the closest one}, that captures the representation of error in this paper. While deep IFL (CAE-MLE) is proposed as a discriminative model and in a generic manner, deep IFL leads to promising results even in comparison to GANs and VAEs based approaches. It is shown that deep IFL (CAE-MLE) provides the state-of-the-art results among discriminative, GANs, and VAEs on MNIST that do not use data augmentation or noise reconstruction as the superiority parts. There is just one approach based on GANs for USPS with 2\% difference that shows better results than deep IFL (CAE-MLE).

\section*{Acknowledgment}
We gratefully thank the high performance computing
team of Northern Arizona University. We acknowledge the support of NVIDIA Corporation with the donation of the Quadro P6000 used for this research.

\bibliographystyle{IEEEtran}
\bibliography{bibio}

\end{document}